# Systematic Comparison of the Influence of Different Data Preprocessing Methods on the Performance of Gait Classifications Using Machine Learning

Johannes Burdack[1], Fabian Horst[1*], Sven Giesselbach[2,3], Ibrahim Hassan[1,4], Sabrina Daffner[5] and Wolfgang I. Schöllhorn[1,6]

[1] Department of Training and Movement Science, Institute of Sport Science, Johannes Gutenberg-University, Mainz, Germany, [2] Knowledge Discovery, Fraunhofer-Institute of Intelligent Analysis and Information Systems (IAIS), Sankt Augustin, Germany, [3] Competence Center Machine Learning Rhine-Ruhr (ML2R), Dortmund, Germany, [4] Faculty of Physical Education, Zagazig University, Zagazig, Egypt, [5] Qimoto, Doctors' Surgery for Sport Medicine and Orthopedics, Wiesbaden, Germany, [6] Department of Wushu, School of Martial Arts, Shanghai University of Sport, Shanghai, China

Human movements are characterized by highly non-linear and multi-dimensional interactions within the motor system. Therefore, the future of human movement analysis requires procedures that enhance the classification of movement patterns into relevant groups and support practitioners in their decisions. In this regard, the use of data-driven techniques seems to be particularly suitable to generate classification models. Recently, an increasing emphasis on machine-learning applications has led to a significant contribution, e.g., in increasing the classification performance. In order to ensure the generalizability of the machine-learning models, different data preprocessing steps are usually carried out to process the measured raw data before the classifications. In the past, various methods have been used for each of these preprocessing steps. However, there are hardly any standard procedures or rather systematic comparisons of these different methods and their impact on the classification performance. Therefore, the aim of this analysis is to compare different combinations of commonly applied data preprocessing steps and test their effects on the classification performance of gait patterns. A publicly available dataset on intra-individual changes of gait patterns was used for this analysis. Forty-two healthy participants performed 6 sessions of 15 gait trials for 1 day. For each trial, two force plates recorded the three-dimensional ground reaction forces (GRFs). The data was preprocessed with the following steps: GRF filtering, time derivative, time normalization, data reduction, weight normalization and data scaling. Subsequently, combinations of all methods from each preprocessing step were analyzed by comparing their prediction performance in a six-session classification using Support Vector Machines, Random Forest Classifiers, Multi-Layer Perceptrons, and Convolutional Neural Networks. The results indicate that filtering GRF data and a supervised data reduction (e.g., using Principal Components Analysis) lead to increased prediction performance of the machine-learning classifiers. Interestingly, the weight normalization and the number of data points (above a certain minimum) in the time normalization does not have a substantial effect. In conclusion, the present results provide first domain-specific recommendations for commonly applied data preprocessing methods and might help to build more comparable and more robust classification models based on machine learning that are suitable for a practical application.

## INTRODUCTION

Human movements are characterized by highly non-linear and multi-dimensional interactions within the motor system (Chau, 2001a; Wolf et al., 2006). In this regard, the use of data-driven techniques seems to be particularly suitable to generate predictive and classification models. In recent years, different approaches based on machine-learning techniques such as Artificial Neural Networks (ANNs), Support Vector Machines (SVMs) or Random Forest Classifiers (RFCs) have been suggested in order to support the decision making of practitioners in the field of human movement analysis, e.g., in classifying movement patterns into relevant groups (Schöllhorn, 2004; Figueiredo et al., 2018). Most machine-learning applications in human movements are found in human gait using biomechanical data (Schöllhorn, 2004; Ferber et al., 2016; Figueiredo et al., 2018; Halilaj et al., 2018; Phinyomark et al., 2018). Although it is generally striking that there are more and more promising applications of machine learning in the field of human movement analysis, the applications are very diverse and differ in their objectives, samples and classification tasks. In order to fulfill the application requirements and to ensure the generalizability of the results, a number of stages are usually carried out to process the raw data in classifications using machine learning. Typically, machine-learning classifications of gait patterns consist of a preprocessing and a classification stage (Figueiredo et al., 2018). The preprocessing stage can be distinguished in feature extraction, feature normalization, and feature selection. The classification stage includes cross validation, model building and validation, as well as evaluation. Different methods have been used for each stage and there is no clear consensus on how to proceed in each of these stages. This is particularly the case for the preprocessing stages of the measured raw data before the classification stage, where there are hardly any recommendations, standard procedures or systematic comparisons of different steps within the preprocessing stage and their impact on the classification accuracy (Slijepcevic et al., 2020). The following six steps, for example, can be derived from the preprocessing stage: (1) Ground reaction force (GRF) filtering, (2) time derivative, (3) time normalization, (4) data reduction, (5) weight normalization, and (6) data scaling.

(1) There are a number of possible noise sources in the recording of biomechanical data. Noise can be reduced by careful experimental procedures, however, cannot be completely removed (Challis, 1999). So far there is less known about optimal filter-cut-off frequencies in biomechanical gait analysis (Schreven et al., 2015). Apart from a limited certainty about an optimal range of filter cut-off frequencies of the individual GRF components, the effect of GRF filtering on the prediction performance of machine-learning classifications has not been reported.

(2) In the majority of cases, time-continuous waveforms or time-discrete gait variables are measured and used for the classification (Schöllhorn, 2004; Figueiredo et al., 2018). Although, some authors also used time derivatives or data in the frequency or frequency-time domain from time-continuous waveforms (Schöllhorn, 2004; Figueiredo et al., 2018). A transformation, which has barely been applied so far, is the first-time derivative of the acceleration, also known as jerk ($\Delta$tGRF) (Flash and Hogan, 1985). However, $\Delta$tGRF might describe human gait more precisely than velocity and acceleration, especially when the GRF is measured. $\Delta$tGRF can be determined directly by calculating the first-time derivative of the GRF measured by force plates.

(3) Feature normalization has been applied in order to achieve more robust classification models (Figueiredo et al., 2018). A normalization in time is commonly applied to normalize the biomechanical waveforms as percentage of the step, stride or stance phase (Kaczmarczyk et al., 2009; Alaqtash et al., 2011a,b; Eskofier et al., 2013; Zhang et al., 2014). It is differentiated among other things between 101 points in time (Eskofier et al., 2013), 1000 points in time (Slijepcevic et al., 2017) or the percentage occurrence per step cycle (Su and Wu, 2000).

(4) The purpose of data reduction is to reduce the amount of data to the most relevant features. A dimensionally reduction is often performed in order to determine which data is to be retained and which can be discarded. The use of dimension reduction can speed up computing time or reduce storage costs for data analysis. However, it should be noted that these feature selection approaches can not only reduce computation costs, but could also improve the classification accuracy (Phinyomark et al., 2018). Beside the unsupervised selection of single time-discrete gait variables (Schöllhorn, 2004; Begg and Kamruzzaman, 2005), typical methods for reducing the dimensionality of the data is, for example, the Principal Component Analysis (Deluzio and Astephen, 2007; Lee et al., 2009; Eskofier et al., 2013; Badesa et al., 2014).

(5) Another way of feature normalization is weight or height normalization. Weight and height normalizations in amplitude are a frequently used method to control for inter-individual differences in kinetic and kinematic variables (Wannop et al., 2012). To what extent the multiplication by a constant factor influences the classification has not yet been investigated to the best of our knowledge.

(6) A third way of feature normalization is data scaling. Data scaling is often performed to normalize the amplitude of one or different variable time courses (Mao et al., 2008; Laroche et al., 2014). The z-score method is mainly used (Begg and Kamruzzaman, 2005; Begg et al., 2005). In machine learning, scaling to a variable or variable waveform the interval [0, 1] or [-1, 1] is common in order to minimize amplitude-related weightings when training the classifiers (Hsu et al., 2003). To the best of our knowledge, it has not yet been investigated whether it makes a difference to scale over a single gait trial or over all trials of one subject in one session.

In summary, there is a lack of domain-specific standard procedures and recommendations, especially for the various data preprocessing steps commonly applied before machine-learning classifications. Therefore, the aim of this analysis is to compare different commonly applied data preprocessing steps and examine their effect on the classification performance using different machine-learning classifiers (ANN, SVM, RFC). A systematic comparison is of particular interest for deriving domain-specific recommendations, finding best practice models and the optimization of machine-learning classifications of human gait data. The analysis is based on the classification problem described by Horst et al. (2017), who investigated intra-individual gait patterns across different time-scales over 1 day.

## MATERIALS AND METHODS

### Sample and Experimental Protocol

The publicly available dataset on intra-individual changes of gait patterns by Horst et al. (2017, 2019a) and two unpublished datasets (Daffner, 2018; Hassan, 2019) following the same experimental protocol were used for this analysis. In total, the joint dataset consisted of 42 physically active participants (22 females, 20 males; 25.6 ± 6.1 years; 1.72 ± 0.09 m; 66.9 ± 10.7 kg) without gait pathology and free of lower extremity injuries. The study was conducted in accordance with the Declaration of Helsinki and all participants were informed of the experimental protocol and provided their written consent. The approval of the ethics committee of the Rhineland-Palatinate Medical Association in Mainz has been received.

As presented in **Figure 1**, the participants performed 6 sessions (S1–S6) of 15 gait trials in each session, while there was no intervention between the sessions. After the first, third and fifth session, the participants had a break of 10 min until the beginning of the subsequent session. Between S2 and S3 was a break of 30 min and between S4 and S5 the break was 90 min. The participants were instructed to walk a 10 m-long path at a self-selected speed barefooted. For each trial, three-dimensional GRFs were recorded by means of two Kistler force plates of type 9287CA (Kistler, Switzerland) at a frequency of 1,000 Hz. The Qualisys Track Manager 2.7 software (Qualisys AB, Sweden) managed the recording. During the investigation, the laboratory environment was kept constant and each subject was analyzed by the same assessor only. Before the first session, each participant carried out 20 familiarization trials to get used to the experimental setup and to determine a starting point for a walk across the force plates. Before each of the following sessions, five familiarization trials were carried out to take into account an effect of practice and to control the individual starting position. In addition, the participants were instructed to look toward a neutral symbol (smiley) on the opposite wall of the laboratory to direct their attention away from targeting the force plates and ensure a natural gait with upright posture. The description of the experimental procedure can be found as well in the original study (Horst et al., 2017).

### Data Preprocessing

The stance phase of the right and left foot was determined using a vertical GRF threshold of 20 Newton. Different combinations of commonly used data preprocessing steps, which typically precede machine-learning classifications of biomechanical gait patterns have been compared (**Figure 2**). Within the introduced stage of preprocessing, the following six data preprocessing steps were investigated: (1. GRF filtering) comparing filtered and unfiltered GRF data. The method described by Challis (1999) was used to determine the optimal cut-off frequencies ($f_c$) for the respective gait trials. The optimal filter frequencies were calculated for each foot and each of the three dimensions in each gait trial separately. (2. Time derivative) comparing the recorded GRF and ΔtGRF, the first-time derivative of the GRF. ΔtGRF was calculated by temporally derivating the GRF for each time interval. (3. Time normalization) comparing the number of time points for the time normalization to the stance phase. Each variable was time normalized to 11, 101 and 1,001 data points, respectively. (4. Data reduction) comparing non-reduced, time-continuous waveforms (TC), time-discrete gait variables (TD) and principle components by a reduction using Principal Component Analysis (PCA) applied to the time-continuous waveforms. The PCA (Hotelling, 1933) is a statistical procedure that uses an orthogonal transformation from a set of observations of potentially correlated variables into a set of values of linearly uncorrelated variables, the so called "principal components." In this transformation, the first principal component explains the largest possible part of the variance. Each subsequent principal component again explains the largest part of the remaining variance, with the restriction that subsequent principal components are orthogonal to the preceding principal components. In our experiment, the resulting features, i.e., the principle components explaining 98% of the total variance, were used as input feature vectors for the classification. The time-discrete gait variables of the fore-aft and medio-lateral shear force were the minimum and the maximum values as well as their occurrence during the stance phase, and of the vertical force the minimum and the two local maxima values as well as their occurrence during the stance phase. This resulted in 28 time-discrete gait variables for GRF data and 24 time-discrete gait variables for $\Delta$tGRF data. (5. Weight normalization) comparing whether weight normalization to the body weight of every session was performed or not. The normalization to the body weight before every season would exclude the impact of any changes in the body mass during the investigation. (6. Data scaling) comparing different data scaling techniques. Scaling is a common procedure for data processing prior to classifications of gait data (Chau, 2001a,b). It was carried out to ensure an equal contribution of all variabilities to the prediction performance and to avoid dominance of variables with greater numeric range (Hsu et al., 2003). On the one hand, this involved a z-transformation over all trials and one over each single trial combined with a scaling to the range of [−1, 1] (Hsu et al., 2003), determined over all trials or over each single trial. The combination of these amplitude normalization methods result in four different scaling methods.

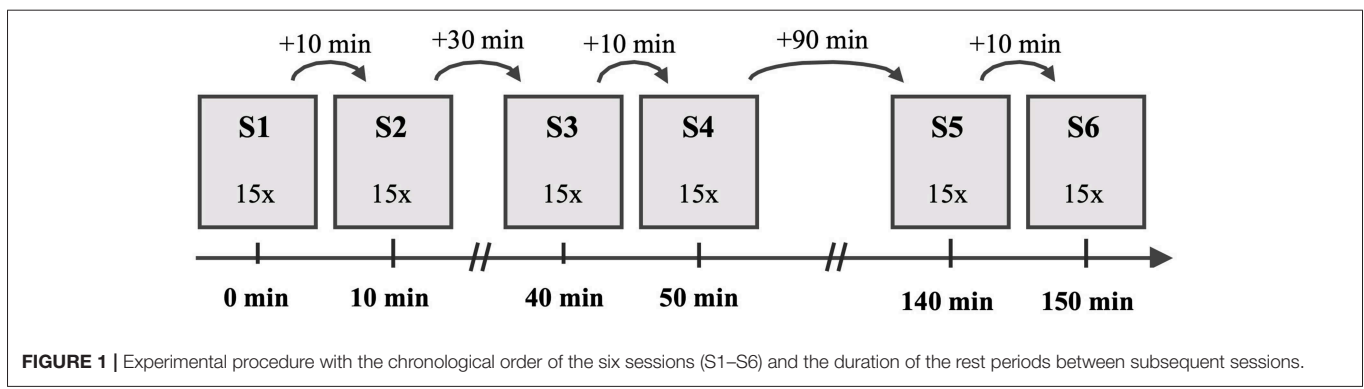

**FIGURE 1** | Experimental procedure with the chronological order of the six sessions (S1–S6) and the duration of the rest periods between subsequent sessions.

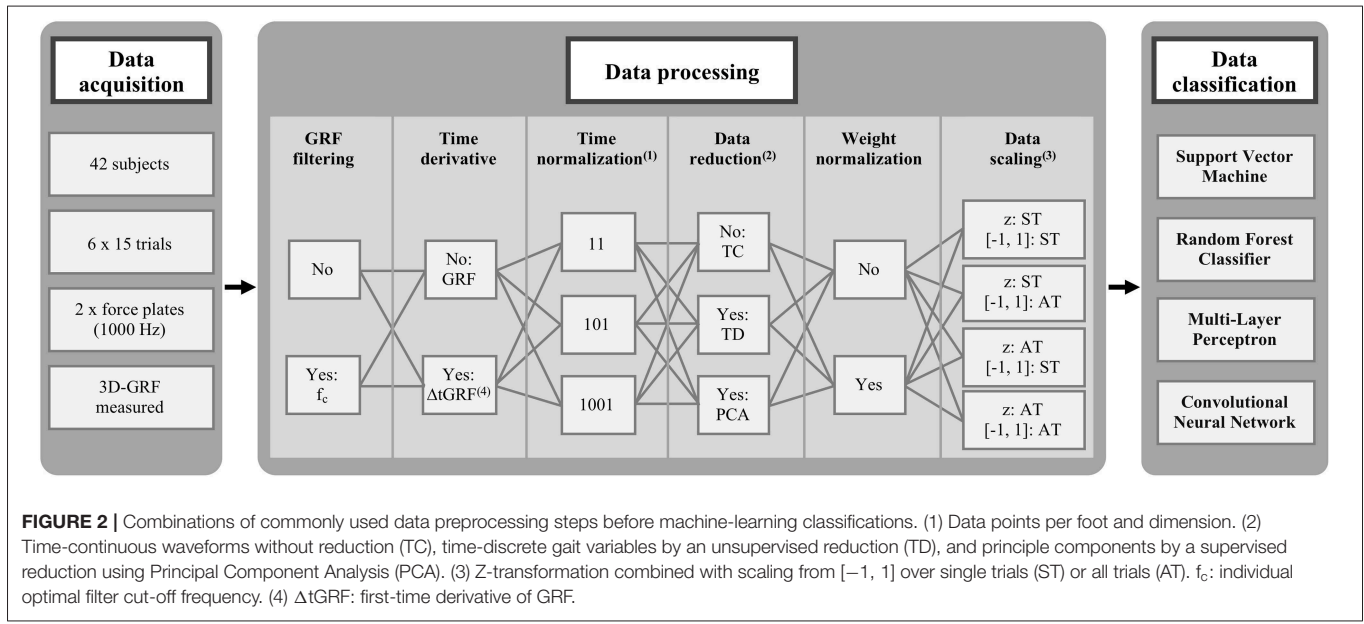

FIGURE 2 | Combinations of commonly used data preprocessing steps before machine-learning classifications. (1) Data points per foot and dimension. (2) Time-continuous waveforms without reduction (TC), time-discrete gait variables by an unsupervised reduction (TD), and principle components by a supervised reduction using Principal Component Analysis (PCA). (3) Z-transformation combined with scaling from [−1, 1] over single trials (ST) or all trials (AT). $f_c$: individual optimal filter cut-off frequency. (4) $\Delta$tGRF: first-time derivative of GRF.

The data preprocessing was managed within Matlab R2017b (MathWorks, USA) and all combinations of each methods of each data preprocessing and classification step were performed in the current analysis in the order described in **Figure 2**. In total, the analysis included 1,152 different combinations of data preprocessing and classification step methods (1,152 = 2 GRF filtering * 2 Time derivative * 3 Time normalization * 3 Data reduction * 2 Weight normalization * 4 Data scaling * 4 Classifier). In the two methods TD and PCA for data reduction, the data scaling could not be applied for all methods. In many cases, all values of a time-discrete gait variable or a principle component were identical [**Figure 2**: Data Scaling z: ST or [−1, 1]: ST] and thus no variance occurred, which is necessary for the calculation of the data scaling. Only, the data scaling over all trials from one subject [**Figure 2**: Data scaling: z: AT, [−1, 1]: AT] could be performed for all three methods of data reduction. In order to keep the number of considered combinations the same for all methods of a data preprocessing step, only the data scaling of all attempts of one subject [**Figure 2**: Data scaling: z: AT, [−1, 1]: AT] was considered for the descriptive and statistical analysis in the results section. This scaling also led to by far the best performance scores. Consequently, 288 different combinations of data preprocessing and classification step methods (288 = 2 GRF filtering * 2 Time derivative * 3 Time normalization * 3 Data reduction * 2 Weight normalization * 1 Data scaling * 4 Classifier) were compared quantitatively with each other on basis of the performance scores.

**TABLE 1** | Length of the resulting input feature vectors depending on different combinations of preprocessing methods.

| Data reduction | Time normalization | Time derivative | GRF filtering | Length of input feature vector |
|---|---|---|---|---|
| TC | 11 | GRF; ΔtGRF | No; Yes | 66 = 11 * 3 * 2 |
|    | 101 | GRF; ΔtGRF | No; Yes | 606 = 101 * 3 * 2 |
|    | 1,001 | GRF; ΔtGRF | No; Yes | 6006 = 1001 * 3 * 2 |
| TD | 11; 101; 1,001 | GRF | No; Yes | 28 = 7 * 2 * 2 |
|    |    | ΔtGRF | No; Yes | 24 = 6 * 2 * 2 |
| PCA | 11 | GRF | No | 46 (44, 47) |
|    |    |    | Yes | 47 (44, 48) |
|    |    | ΔtGRF | No | 53 (49, 55) |
|    |    |    | Yes | 38 (43, 46) |
|    | 101 | GRF | No | 78 (73, 83) |
|    |    |    | Yes | 72.5 (69, 79) |
|    |    | ΔtGRF | No | 239 (210, 268) |
|    |    |    | Yes | 108 (97, 119) |
|    | 1,001 | GRF | No | 79 (73, 84) |
|    |    |    | Yes | 72 (68, 79) |
|    |    | ΔtGRF | No | 369 (341, 386) |
|    |    |    | Yes | 108 (97, 120) |

*TC, time-continuous waveforms for three dimensions (\*3) and two steps (\*2); TD, time-discrete gait variables of minima and maxima of the three dimensions (GRF: 7; ΔtGRF: 6) for two steps (\*2) and their relative occurrences (\*2); PCA, Median and interquartile distance of the number of principle components.*

## Data Classification

The intra-individual classification of gait patterns was based on the 90 gait trials (90 = 6 sessions × 15 trials) of each participant. For each trial, a concatenated vector of the three-dimensional variables of both force plates was used for the classification. Due to the different time normalization and data reduction methods, the resulting length of the input feature vectors differed (**Table 1**).

The classification based on the following four supervised machine-learning classifiers with an exhaustive hyper-parameter search: (1) Support Vector Machines (SVMs) (Boser et al., 1992; Cortes and Vapnik, 1995; Müller et al., 2001; Schölkopf and Smola, 2002) using a linear kernel and a grid search to determine the best cost parameter $C = 2^{-5}, 2^{-4.75}, \ldots, 2^{15}$. (2) Random Forest Classifiers (RFCs) (Breiman, 2001) with the Gini coefficient as decision criterion. Different numbers of trees (n_estimators = 200, 225, ..., 350) and maximal tree depth (n_depth = 4, 5, ..., 8) were determined empirically via grid search. (3) Multi-Layer Perceptrons (MLPs) (Bishop, 1995) with one hidden layer of size $2^6$ (= 64 neurons) and 2,000 iterations with the weight optimization algorithm Adam ($\beta 1 = 0.9$, $\beta 2 = 0.999$, $\varepsilon = 10^{-8}$). The learning rate regularization parameter $\alpha$ (= $10^{-1}, 10^{-2}, \ldots, 10^{-7}$) was determined via grid search in the cross-validation. (4) Convolutional Neural Networks (CNNs) (LeCun et al., 2015) consisting of three convolutional layers and one fully connected layer. The first convolutional layers contained 24 filters with a kernel size of 8, a stride of 2 and a padding of 4. The second contained 32 filters with a kernel size of 8, a stride of 2 and a padding of 4. The third convolutional layer contained 48 filters with a kernel size of 6, a stride of 3 and a padding of 3. After each convolutional layer a ReLU activation was performed and after a fully connected layer a SoftMax was used to obtain probability of each of the classes. This architecture follows CNNs previously used for the classification of GRF data (Horst et al., 2019b). The ability to distinguish gait patterns of one test session from gait patterns of other test sessions was investigated in a multi-class classification (six-session classification) setting. For the evaluation of the prediction performance, the F1-, precision- and recall-scores were calculated over a stratified 15-fold cross validation configuration. 78 of 90 parts of the data were used for training, 6 of 90 parts were used as a validation set and the remaining 6 of 90 parts was reserved for testing. The 6 samples per test split were evenly distributed across all session partitions and are excluded from the complete training and validation process. Only 6 samples were selected for the test split because we wanted to guarantee as much training data as possible. In order to get meaningful results, the Training Validation Test splitting was stratified repeated 15 times so that each of the 90 gait trials was exactly once in the test set. The classification was performed within Python 3.6.3 (Python Software Foundation, USA) using the scikit-learn toolbox (0.19.2) (Pedregosa et al., 2011) and PyTorch (1.2.0) (Paszke et al., 2019).

The evaluation was carried out by calculating the performance indicators (accuracy, F1-score, precision and recall) defined by the number of true positives (TP), true negatives (TN), false positives (FP), and false negatives (FN):

$$Accuracy = \frac{TP + TN}{TP + TN + FP + FN}$$

$$Precision = \frac{TP}{TP + FP}$$

$$Recall = \frac{TP}{TP + FN}$$

$$F1 - score = 2 * \frac{Precision * Recall}{Precision + Recall}$$

Please note that since this is a balanced data set for multi-class classification, the accuracy corresponds exactly to the recall.

## Statistical Analysis

For the comparison of the different combinations of the described preprocessing steps, the mean performance scores were compared statistically. Each mean value combined all combinations of preprocessing steps where the preprocessing method was part of. The Shapiro-Wilk test showed that none of the examined variables violated the normal distribution assumption ($p \geq 0.109$). For the comparison of all combinations of the preprocessing methods, paired-samples *t*-test and repeated-measures ANOVAs were calculated for the variables of time derivative, GRF filtering and weight normalization. For the ANOVAs *post hoc* Bonferroni corrected paired-samples *t*-tests were calculated for the variables of time normalization, data reduction and classifier. Furthermore, the effect sizes d and $\eta_p^2$ were calculated; d and $\eta_p^2$ are considered a small effect for $|d| = 0.2$ and $\eta_p^2 < 0.06$, a medium effect for $|d| = 0.5$ and

TABLE 2 | Mean F1-score for each individual participant depending on each preprocessing method and machine-learning classifier.

| | GRF filtering | | Time derivative | | Time normalization | | | Data reduction | | | Weight normalization | | Machine-learning classifier | | | |
|---|---|---|---|---|---|---|---|---|---|---|---|---|---|---|---|---|
| | No | Yes | GRF | ΔtGRF | 11 | 101 | 1001 | TC | TD | PCA | No | Yes | SVM | RFC | MLP | CNN |
| S01 | 38.3 | 42.0 | 41.4 | 39.0 | 34.9 | 42.4 | 43.2 | 40.2 | 25.7 | 54.5 | 39.9 | 40.4 | 46.4 | 45.0 | 36.3 | 32.9 |
| S02 | 24.5 | 29.4 | 27.9 | 26.0 | 23.2 | 28.4 | 29.2 | 24.5 | 20.6 | 35.7 | 27.2 | 26.7 | 30.0 | 30.7 | 23.9 | 23.2 |
| S03 | 36.9 | 43.5 | 40.6 | 39.8 | 36.1 | 44.3 | 40.2 | 40.9 | 27.6 | 52.1 | 40.1 | 40.3 | 44.8 | 43.9 | 38.2 | 33.9 |
| S04 | 42.9 | 50.0 | 48.2 | 45.1 | 41.5 | 48.8 | 49.5 | 48.8 | 36.5 | 53.6 | 45.9 | 47.3 | 51.4 | 56.2 | 42.0 | 36.8 |
| S05 | 49.9 | 50.2 | 52.0 | 48.1 | 47.1 | 51.1 | 52.0 | 50.2 | 36.5 | 63.6 | 49.6 | 50.6 | 56.7 | 56.2 | 45.5 | 41.9 |
| S06 | 38.4 | 39.8 | 39.3 | 38.9 | 32.0 | 42.3 | 43.2 | 38.9 | 28.0 | 49.8 | 39.3 | 38.9 | 42.8 | 44.6 | 38.2 | 30.8 |
| S07 | 31.5 | 40.5 | 35.4 | 36.7 | 30.5 | 40.0 | 37.6 | 34.6 | 28.7 | 44.8 | 36.2 | 35.9 | 39.2 | 41.7 | 32.8 | 30.3 |
| S08 | 42.7 | 49.0 | 46.4 | 45.4 | 41.4 | 49.0 | 47.0 | 47.3 | 38.2 | 51.8 | 45.7 | 46.0 | 49.0 | 52.1 | 44.9 | 37.6 |
| S09 | 43.2 | 47.2 | 46.1 | 44.3 | 39.1 | 49.8 | 46.7 | 43.5 | 34.0 | 58.1 | 45.3 | 45.1 | 51.2 | 49.7 | 41.2 | 38.7 |
| S10 | 41.3 | 40.3 | 41.2 | 40.4 | 34.2 | 44.1 | 44.1 | 44.4 | 27.5 | 50.5 | 40.5 | 41.1 | 43.8 | 43.2 | 42.2 | 33.9 |
| S11 | 38.5 | 40.7 | 42.5 | 36.7 | 35.3 | 42.8 | 40.8 | 42.0 | 27.6 | 49.3 | 39.5 | 39.7 | 44.0 | 45.1 | 35.2 | 34.2 |
| S12 | 34.1 | 31.9 | 36.2 | 29.8 | 27.9 | 35.4 | 35.7 | 35.3 | 22.9 | 40.9 | 33.5 | 32.6 | 36.7 | 34.9 | 34.1 | 26.3 |
| S13 | 31.7 | 34.5 | 34.4 | 31.8 | 28.6 | 36.9 | 33.8 | 32.5 | 27.8 | 39.0 | 32.8 | 33.4 | 36.9 | 36.6 | 31.2 | 27.6 |
| S14 | 33.9 | 34.0 | 38.1 | 29.8 | 28.3 | 37.3 | 36.2 | 35.7 | 24.4 | 41.7 | 34.1 | 33.8 | 36.3 | 35.4 | 35.7 | 28.3 |
| S15 | 39.9 | 45.3 | 46.8 | 38.4 | 36.8 | 46.7 | 44.2 | 42.5 | 31.2 | 54.0 | 42.8 | 42.3 | 48.7 | 46.2 | 39.5 | 35.8 |
| S16 | 32.0 | 32.9 | 32.9 | 31.9 | 27.5 | 34.5 | 35.3 | 33.8 | 23.3 | 40.2 | 32.7 | 32.2 | 34.3 | 34.6 | 34.6 | 26.3 |
| S17 | 29.3 | 30.0 | 31.7 | 27.6 | 22.6 | 33.4 | 32.6 | 30.0 | 21.7 | 36.9 | 29.7 | 29.7 | 33.0 | 31.5 | 29.4 | 24.8 |
| S18 | 24.4 | 26.9 | 25.6 | 25.8 | 22.8 | 27.0 | 27.3 | 28.3 | 17.9 | 30.9 | 25.7 | 25.7 | 27.6 | 26.7 | 27.5 | 21.0 |
| S19 | 27.3 | 28.7 | 31.5 | 24.5 | 25.0 | 29.4 | 29.5 | 26.8 | 22.7 | 34.5 | 27.9 | 28.0 | 31.0 | 30.6 | 25.5 | 24.8 |
| S20 | 29.3 | 34.0 | 32.4 | 31.0 | 26.6 | 33.7 | 34.7 | 31.9 | 25.7 | 37.4 | 31.8 | 31.6 | 34.4 | 36.3 | 30.0 | 26.2 |
| S21 | 27.7 | 29.6 | 30.9 | 26.4 | 25.4 | 30.8 | 29.7 | 28.1 | 22.4 | 35.2 | 28.6 | 28.8 | 31.2 | 33.1 | 26.4 | 24.0 |
| S22 | 32.3 | 33.6 | 36.4 | 29.5 | 28.6 | 34.5 | 35.7 | 34.1 | 24.2 | 40.5 | 33.2 | 32.7 | 33.6 | 35.3 | 35.6 | 27.2 |
| S23 | 31.7 | 35.0 | 34.6 | 32.1 | 28.5 | 35.0 | 36.5 | 33.8 | 25.6 | 40.7 | 33.1 | 33.6 | 34.9 | 39.0 | 32.8 | 26.6 |
| S24 | 35.4 | 43.3 | 40.1 | 38.6 | 33.9 | 41.3 | 42.9 | 39.7 | 32.1 | 46.3 | 39.7 | 39.1 | 42.8 | 43.8 | 39.2 | 31.6 |
| S25 | 34.7 | 41.9 | 39.3 | 37.4 | 34.7 | 41.6 | 38.7 | 37.1 | 33.0 | 44.8 | 38.3 | 38.4 | 40.2 | 43.8 | 36.8 | 32.5 |
| S26 | 47.6 | 49.9 | 53.6 | 43.8 | 42.3 | 51.3 | 52.5 | 52.2 | 41.7 | 52.1 | 48.5 | 48.9 | 52.9 | 56.0 | 47.0 | 38.6 |
| S27 | 31.5 | 31.8 | 30.4 | 32.8 | 26.5 | 34.2 | 34.2 | 33.6 | 24.6 | 36.7 | 31.6 | 31.6 | 32.8 | 35.2 | 31.5 | 27.0 |
| S28 | 35.9 | 45.5 | 41.9 | 39.4 | 33.5 | 45.1 | 43.5 | 43.0 | 29.9 | 49.2 | 40.9 | 40.5 | 43.5 | 43.9 | 42.3 | 33.0 |
| S29 | 32.2 | 36.1 | 33.1 | 35.2 | 30.1 | 36.5 | 35.8 | 36.3 | 22.6 | 43.6 | 34.5 | 33.9 | 36.8 | 35.7 | 36.0 | 28.1 |
| S30 | 31.1 | 33.1 | 35.4 | 28.9 | 28.3 | 32.6 | 35.4 | 35.1 | 21.6 | 39.0 | 31.9 | 32.3 | 33.4 | 37.1 | 32.5 | 25.4 |
| S31 | 51.3 | 53.7 | 54.5 | 50.5 | 44.5 | 56.0 | 57.0 | 58.6 | 36.6 | 62.2 | 52.4 | 52.5 | 56.8 | 58.3 | 53.5 | 41.3 |
| S32 | 43.0 | 45.9 | 47.4 | 41.5 | 38.6 | 46.0 | 48.7 | 49.5 | 31.1 | 52.7 | 44.7 | 44.2 | 47.9 | 50.3 | 44.0 | 35.5 |
| S33 | 35.7 | 41.4 | 39.7 | 37.4 | 32.1 | 40.4 | 43.1 | 41.6 | 23.5 | 50.3 | 38.2 | 38.9 | 42.7 | 41.6 | 39.2 | 30.7 |
| S34 | 49.8 | 51.8 | 53.8 | 47.8 | 44.5 | 53.2 | 54.7 | 52.1 | 39.0 | 61.4 | 50.7 | 50.9 | 54.1 | 57.5 | 51.3 | 40.4 |
| S35 | 38.4 | 45.4 | 45.3 | 38.8 | 35.1 | 45.5 | 45.5 | 45.6 | 25.7 | 53.7 | 42.2 | 41.8 | 45.8 | 47.7 | 42.2 | 32.2 |
| S36 | 36.9 | 39.3 | 41.0 | 35.3 | 32.9 | 40.7 | 40.7 | 39.5 | 29.8 | 45.1 | 37.9 | 38.3 | 41.2 | 43.1 | 36.8 | 31.3 |
| S37 | 30.9 | 33.7 | 35.8 | 28.9 | 27.7 | 33.5 | 35.8 | 33.8 | 20.3 | 42.9 | 32.2 | 32.5 | 35.3 | 33.3 | 34.3 | 26.4 |
| S38 | 35.1 | 38.2 | 39.0 | 34.3 | 30.9 | 38.7 | 40.3 | 37.8 | 26.5 | 45.6 | 36.7 | 36.6 | 39.5 | 40.9 | 37.1 | 29.1 |
| S39 | 41.6 | 43.2 | 46.1 | 38.7 | 39.0 | 43.1 | 45.1 | 47.4 | 28.1 | 51.7 | 42.4 | 42.4 | 44.3 | 48.6 | 42.8 | 33.9 |
| S40 | 41.4 | 48.9 | 48.8 | 41.5 | 37.1 | 47.9 | 50.4 | 48.1 | 30.3 | 56.9 | 45.1 | 45.2 | 48.9 | 50.5 | 46.8 | 34.4 |
| S41 | 38.4 | 43.2 | 43.9 | 37.6 | 34.7 | 42.9 | 44.7 | 44.4 | 28.1 | 49.7 | 40.6 | 41.0 | 42.6 | 48.3 | 40.5 | 31.7 |
| S42 | 27.2 | 29.4 | 31.3 | 25.4 | 25.5 | 28.2 | 31.4 | 29.7 | 21.6 | 33.0 | 28.3 | 28.3 | 29.2 | 31.4 | 28.3 | 24.3 |
| M | 36.2 | 39.6 | 39.8 | 36.0 | 32.8 | 40.4 | 40.6 | 39.4 | 27.8 | 46.5 | 37.9 | 37.9 | 41.2 | 42.3 | 37.3 | 31.0 |
| SD | 6.7 | 7.3 | 7.3 | 6.8 | 6.3 | 7.3 | 7.3 | 7.8 | 5.6 | 8.3 | 6.8 | 7.0 | 7.8 | 8.3 | 6.8 | 5.3 |

The mean precision and mean recall (= accuracy) scores for each individual participant depending on each preprocessing method and machine-learning classifier can be found in **Supplementary Tables S1, S2**.
Each mean value combines all combinations of preprocessing steps where the preprocessing method was part of (n = 42).

$0.06 < \eta_p^2 < 0.14$ and a large effect for $|d| = 0.8$ and $\eta_p^2 > 0.14$ (Cohen, 1988). The *p*-value at which research is considered worth to be continued (Fisher, 1922) has been set to $p = 0.05$. To determine a best practice model, all combinations of data preprocessing methods were ranked according to their mean performance scores over 15-fold cross validation and the rank sum was calculated.

# RESULTS

## Average Performance of Different Data Preprocessing Methods

The analysis compares 288 different combinations of data preprocessing methods based on the resulting F1-score. **Table 2** displays the mean F1-score for each individual participant over the 15-fold cross validation (**Supplementary Tables S1**, **S2** show the mean precision and recall values).

**Figure 3** shows the mean F1-scores over all participants. It is noticeable that the highest mean F1-scores were achieved using PCA, while the normalization to 101 and 1,001 data points or the weighting has only a minor effect on the F1-score. The time normalization to only 11 data points and the reduction to time-discrete gait variables gave particularly low mean classification scores. Concerning the machine-learning classifiers, the RFCs achieved the highest mean F1-scores followed by the SVMs, MLPs, and CNNs.

## GRF Filtering

A paired-samples *t*-test was performed to determine if there were differences in F1-score in unfiltered GRF data compared to $f_c$–filtered GRF data across all participants. The mean F1-score of the filtered GRF data ($M = 39.6\%$, $SD = 7.3\%$) was significantly higher than that of the unfiltered GRF data ($M = 36.2\%$, $SD = 6.7\%$). The effect size, however, was small $[t_{(41)} = 8.200, p < 0.001, |d| = 0.492]$.

## Time Derivative

A paired-samples *t*-test was conducted to compare the F1-score of GRF and $\Delta$tGRF across all participants. The mean F1-score of GRF ($M = 39.8\%$, $SD = 7.3\%$) was significantly higher than that of $\Delta$tGRF ($M = 36.0\%$, $SD = 6.8\%$) and showed a medium effect size $[t_{(41)} = 8.162, p < 0.001, |d| = 0.540]$.

## Time Normalization

A repeated-measures ANOVA determined that there is a significant global effect with large effect size of F1-score between time normalization to 11, 101 and 1,001 data points $[F_{(2.000, 82.000)} = 367.115, p < 0.001, \eta_p^2 = 0.900]$. *Post hoc* paired-samples *t*-test with Bonferroni correction revealed that there is no significant difference $[t_{(41)} = -0.741, p = 0.463, |d| = 0.031]$ between a time normalization to 101 ($M = 40.4\%$, $SD = 7.3\%$) data points and 1,001 data points ($M = 40.6\%$, $SD = 7.3\%$). However, the time normalization to 101 data points performed significantly better $[t_{(41)} = 22.397, p < 0.001, |d| = 1.118]$ than time normalized to 11 data points ($M = 32.8\%$, $SD = 6.3\%$). Also the time normalization to 1,001 data points performed significantly better than to 11 data points $[t_{(41)} = 21.789, p < 0.001, |d| = 1.150]$. Both effect sizes are considered as large.

## Data Reduction

A one-way repeated-measures ANOVA was conducted to compare the F1-scores of PCA ($M = 54.9\%$, $SD = 8.5\%$), TC ($M = 50.9\%$, $SD = 8.8\%$), and TD ($M = 37.5\%$, $SD = 6.5\%$). The Huynh-Feldt corrected results showed a highly significant main effect with a large effect size $[F_{(1.594, 65.365)} = 378.372, p < 0.001, \eta_p^2 = 0.902]$. Bonferroni corrected *post hoc* paired-samples *t*-tests showed that PCA performed significantly better than TC $[t_{(41)} = 14.540, p < 0.001, |d| = 0.884]$ and TD $[t_{(41)} = 22.658, p < 0.001, |d| = 2.635]$. The effect size for both comparisons is considered as large. Furthermore, TC performed also significantly better than TD with a large effect size $[t_{(41)} = 16.516, p < 0.001, |d| = 1.701]$.

## Weight Normalization

A paired-samples *t*-test was conducted to compare the F1-scores of weight-normalized and non-weight-normalized data across all participants. There was no significant difference $[t_{(41)} = -0.644, p = 0.523, |d| = 0.006]$ in the F1-scores for non-weight-normalized data ($M = 37.9\%$, $SD = 6.8\%$) and weight-normalized data ($M = 37.9\%$, $SD = 7.0\%$).

## Machine-Learning Classifier

A repeated-measures ANOVA with Huynh-Feldt correction showed a highly significantly global effect with large effect size $[F_{(1.130, 103.478)} = 240.138, p < 0.001, \eta_p^2 = 0.854]$ between the predicted F1-scores by the SVMs ($M = 41.2\%$, $SD = 7.8\%$), RFCs ($M = 42.3\%$, $SD = 8.3\%$), MLPs ($M = 37.3\%$, $SD = 6.8\%$), and CNNs ($M = 31.0\%$, $SD = 5.3\%$). *Post hoc* Bonferroni corrected paired-samples *t*-test revealed that the RFCs performed significantly better, with a small effect size, than the SVMs $[t_{(41)} = 3.531, p = 0.001, |d| = 0.140]$, with a medium effect size than the MLPs $[t_{(41)} = 9.459, p < 0.001, |d| = 0.664]$ and with a large effect size than the CNNs $[t_{(41)} = 20.780, p < 0.001, |d| = 1.625]$. Also the SVMs performed significantly better than the MLPs with a medium effect $[t_{(41)} = 8.115, p < 0.001, |d| = 0.534]$ and significantly better than the CNNs with a large effect $[t_{(41)} = 23.811, p < 0.001, |d| = 1.530]$. Furthermore, the MLPs performed significantly better than the CNNs with a large effect $[t_{(41)} = 13.725, p < 0.001, |d| = 1.035]$.

## Best Practice Combinations of Different Data Preprocessing Methods

In addition to the mean F1-scores for each method of all preprocessing and classification steps, **Table 3** shows the 30 combinations with the highest overall F1-scores, including precision and recall (the complete list including precision and recall can be found in **Supplementary Table S3**). It is particularly noticeable that the first 18 ranks were all achieved using PCA for data reduction. Furthermore, the first eight ranked combinations used GRF data. The first twelve ranked combinations were classified with SVMs, while the highest F1-score was 13th with MLP, 27th with RFC and 57th with CNN.

**Table 4** shows the rank scores of all classifications performed for the 288 combinations of the different preprocessing steps according to the F1-scores (**Supplementary Tables S4**, **S5** display the rank score depending on precision and recall). The PCA achieved a particularly high rank score with 87.5% of the maximum achievable rank score. In addition, the GRF with 73.9% and the GRF filtering with 73.4% finished with high rank scores. Again, there are no or only minor differences within the weight normalization and the time normalization to 101 and 1,001 data points. Among the classifiers, the RFCs achieved the highest rank score, just ahead of the SVMs and MLPs and quite far in advance of CNNs.

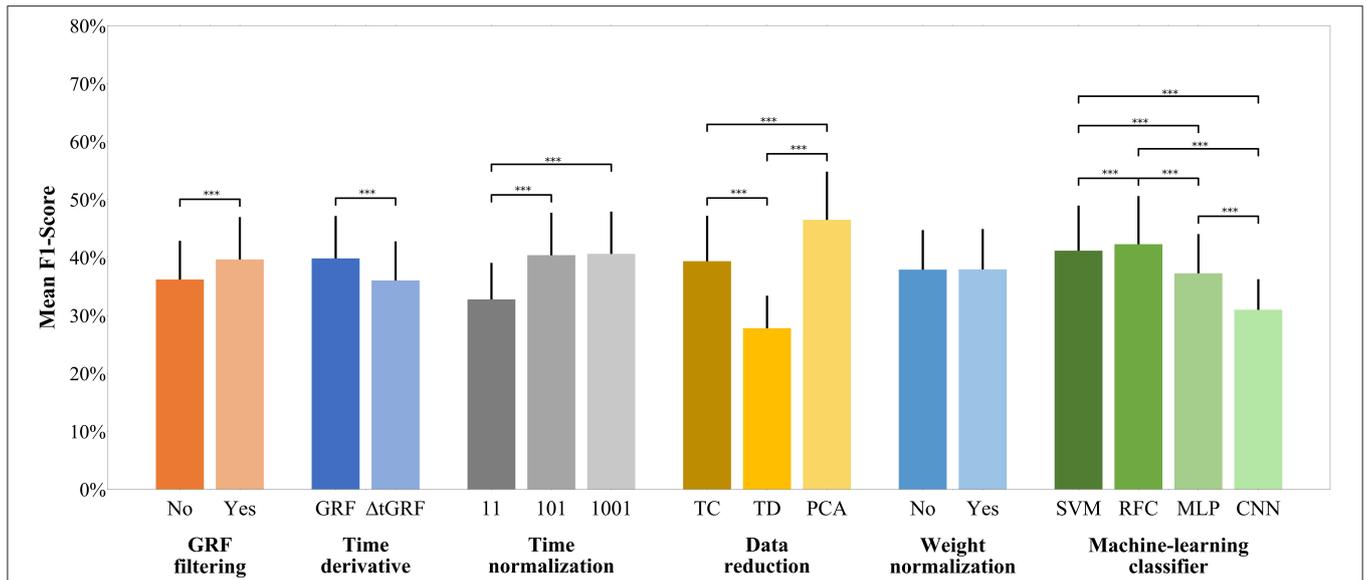

FIGURE 3 | F1-score of each preprocessing step across all participants. The y-axis shows the mean F1-score achieved. The bar charts show the mean value and the standard deviation depending on the respective preprocessing step. The parentheses show a statistically significant effect. Random Baseline = 16.7%; ***$p \leq 0.001$.

## DISCUSSION

A growing number of promising machine-learning applications could be found in the field of human movement analysis. However, these approaches differ in terms of objectives, samples, and classification tasks. Furthermore, there is a lack of standard procedures and recommendations within the different methodological approaches, especially with respect to data preprocessing steps usually performed prior to machine-learning classification. In this regard, the current analysis comprised a systematic comparison of different preprocessing steps and their effects on the prediction performance of different machine-learning classifiers. The results revealed first domain-specific recommendations for the preprocessing of GRF data prior to machine-learning classifications. This includes, for example, benefits of filtering GRF data and supervised data reduction techniques (e.g., PCA) compared to non-reduced (time-continuous waveforms) or unsupervised data reduction techniques (time-discrete gait variables). On the other hand, the results indicate that the normalization to a constant factor (weight normalization) and the number of data points (above a certain minimum) used during time normalization seem to have little influence on the prediction performance. Furthermore, the first-time derivative ($\Delta$tGRF) could not achieve advantages over the GRF in terms of prediction performance.

In general, the present results can help to find domain-specific standard procedures for the preprocessing of data that may enable to improve machine-learning classifications in human movement analysis make different approaches better comparable in the future. It should be noted, however, that the results presented are based solely on prediction performance and do not provide information about the effects on the trained models.

### GRF Filtering

The present results indicate that the filtered GRF data led to significantly higher mean F1-scores and rank scores than the unfiltered GRF data. The results were especially striking for the classifications of $\Delta$tGRF data. While no clear trend could be derived for the best-ranked combinations of GRF data, most of the best-ranked combinations of $\Delta$tGRF data were filtered. To our knowledge, this analysis was the first that investigated whether a filter (using an optimal filter cut-off frequency) affects the prediction performance of GRF data in human gait (Schreven et al., 2015). The present findings suggest that machine-learning classification should use filtered GRF data. However, it should be noted that the estimation of the optimal filter cut-off frequency using the method described by Challis (1999) is only one out of several possibilities to set a cut-off frequency. Because the individual filter cut-off frequencies were separately calculated for trial and each variable, so it is not yet possible to recommend a generally valid unique cut-off frequency.

### Time Derivative

With respect to the feature extraction using the first-time derivative, our analysis revealed that the GRF achieved significantly higher F1-scores compared to the $\Delta$tGRF. In

TABLE 3 | Top 30 combinations of preprocessing methods, ranked by the mean F1-score over the 15-fold cross validation ($n = 42$).

| Rank | GRF filtering | Time derivative | Time normalization | Data reduction | Weight normalization | Machine-learning classifier | M | SD |
|---|---|---|---|---|---|---|---|---|
| 1 | No | GRF | 1,001 | PCA | No | SVM | 54.4 | 9.8 |
| 2 | No | GRF | 101 | PCA | Yes | SVM | 54.2 | 10.3 |
| 3 | No | GRF | 1,001 | PCA | Yes | SVM | 54.1 | 11.2 |
| 4 | Yes | GRF | 1,001 | PCA | No | SVM | 54.0 | 10.3 |
| 5 | Yes | GRF | 101 | PCA | No | SVM | 53.9 | 10.3 |
| 6 | No | GRF | 101 | PCA | No | SVM | 53.8 | 9.8 |
| 7 | Yes | GRF | 1,001 | PCA | Yes | SVM | 53.7 | 11.6 |
| 8 | Yes | GRF | 101 | PCA | Yes | SVM | 53.6 | 11.3 |
| 9 | Yes | ΔtGRF | 1,001 | PCA | No | SVM | 53.5 | 10.6 |
| 10 | Yes | ΔtGRF | 101 | PCA | No | SVM | 53.2 | 10.3 |
| 11 | Yes | ΔtGRF | 101 | PCA | Yes | SVM | 53.2 | 10.8 |
| 12 | Yes | ΔtGRF | 1,001 | PCA | Yes | SVM | 53.2 | 10.6 |
| 13 | No | GRF | 1,001 | PCA | No | MLP | 53.0 | 9.7 |
| 14 | Yes | GRF | 101 | PCA | No | MLP | 52.7 | 9.2 |
| 15 | Yes | GRF | 1,001 | PCA | No | MLP | 52.7 | 10.0 |
| 16 | No | GRF | 101 | PCA | Yes | MLP | 52.6 | 10.2 |
| 17 | Yes | GRF | 1,001 | PCA | Yes | MLP | 52.6 | 10.2 |
| 18 | No | GRF | 1,001 | PCA | Yes | MLP | 52.3 | 9.6 |
| 19 | Yes | ΔtGRF | 101 | TC | Yes | RFC | 52.1 | 10.4 |
| 20 | No | GRF | 101 | PCA | No | MLP | 52.1 | 9.3 |
| 21 | Yes | GRF | 101 | PCA | Yes | MLP | 52.1 | 10.5 |
| 22 | Yes | ΔtGRF | 101 | TC | Yes | MLP | 51.6 | 9.6 |
| 23 | Yes | ΔtGRF | 1,001 | PCA | Yes | MLP | 51.6 | 9.4 |
| 24 | Yes | ΔtGRF | 101 | PCA | No | MLP | 51.6 | 10.6 |
| 25 | Yes | ΔtGRF | 101 | TC | No | RFC | 51.5 | 10.8 |
| 26 | Yes | ΔtGRF | 1,001 | PCA | No | MLP | 51.4 | 9.3 |
| 27 | Yes | ΔtGRF | 1,001 | TC | Yes | RFC | 51.4 | 10.7 |
| 28 | Yes | ΔtGRF | 101 | PCA | Yes | MLP | 51.4 | 10.5 |
| 29 | Yes | ΔtGRF | 101 | TC | No | MLP | 51.1 | 9.9 |
| 30 | Yes | ΔtGRF | 1,001 | TC | No | RFC | 51.1 | 10.5 |

(1) The rounded percentage means and standard deviations of the F1-scores are shown; therefore, identical values may occur in the table. However, there are no pairwise identical values, so the ranking is unique. (2) A table including precision and recall (= accuracy) can be found in **Supplementary Table S3**.

addition, the highest prediction F1-scors were also achieved with the GRF. However, it needs to be noted that the highest F1-score using ΔtGRF data were <1% lower than the highest F1-score using GRF data. Because the time derivative alone did not increase the prediction performance, it might be helpful to aggregate different feature extraction methods to improve classification models (Slijepcevic et al., 2020).

## Time Normalization

The time normalization to 101 and 1,001 data points was significantly better than that to only 11 data points. These results are in line with current research, where 101 and 1,001 values are commonly used (Eskofier et al., 2013; Slijepcevic et al., 2017). Three of the four best ranks were achieved using the time normalization to 1,001 data points, but these were only slightly higher than those time normalized to 101 data points. In both methods, the best prediction performances where achieved in combination with PCA. In terms of computational costs, it is advisable to weigh up to what extent relatively small improvements in the prediction performance justify the additional time required for classification. Furthermore, if computational cost is an important factor, a time normalization to fewer data points (above a certain minimum) could also be useful, since the results showed only little influence on the prediction performance.

## Data Reduction

This analysis showed that PCA, which is frequently used in research (Figueiredo et al., 2018; Halilaj et al., 2018; Phinyomark et al., 2018), also achieves the highest F1-scores and ranks, compared with time-continuous waveforms and time-discrete gait variables. The highest F1-score of a machine-learning model based on time-continuous waveforms was 2.3% lower than that of PCA. Machine-learning models solely according to time-discrete characteristics is not recommended based on these analysis results. In line with Phinyomark et al. (2018),

TABLE 4 | Rank scores of all combinations of preprocessing methods depending on their mean F1-score over the 15-fold cross validation ($n = 42$).

| | GRF filtering | | Time derivative | | Time normalization | | | Data reduction | | | Weight normalization | | Classifier | | | |
|---|---|---|---|---|---|---|---|---|---|---|---|---|---|---|---|---|
| | No | Yes | GRF | ΔtGRF | 11 | 101 | 1001 | TC | TD | PCA | No | Yes | SVM | RFC | MLP | CNN |
| Score | 18,564 | 22,764 | 22,953 | 18,375 | 10,392 | 15,512 | 15,424 | 14,337 | 6,870 | 20,121 | 20,635 | 20,693 | 11,926 | 12,170 | 10,373 | 6,859 |
| % max | 39.9 | 60.1 | 61.0 | 39.0 | 21.1 | 39.6 | 39.3 | 35.4 | 8.4 | 56.3 | 49.9 | 50.1 | 30.1 | 30.9 | 25.1 | 13.8 |

(1) The total rank score is for each preprocessing step is 41,328. For GRF filtering, time derivative, and weight normalization the minimum rank score is 10,296 (0.0%) and the maximum rank score is 31,032 (100.0%). For time normalization and data reduction the minimum rank score is 4,560 (0.0%) and the maximum is 22,992 (0.0%) and the maximum is 2,556 (0%). %max: relative rank score of ranks scaled to the interval between the minimum rank score and the maximum total rank score. (2) The rank scores for precision and recall (= accuracy) can be found in Supplementary Tables S4, S5.

reducing the amount of data to the relevant characteristics is not only a cost-reducing method, but can also improve machine-learning classifications.

## Weight Normalization

While weight normalization is necessary in inter-individual comparisons (Mao et al., 2008; Laroche et al., 2014), there have been no recommendations regarding intra-personal comparisons so far. The results of this analysis suggest that performing or not performing weight normalization leads to almost the same results and therefore shows no difference in prediction performance. Consequently, multiplication by a constant factor seems to play no role in the machine-learning classifications. This could be particularly interesting if different datasets are combined.

## Machine-Learning Classifier

Four commonly used machine-learning classifiers (SVM, RFC, MLP, and CNN) were compared in this analysis. The RFCs achieved significantly higher mean F1-scores across all data preprocessing methods than the SVMs, MLPs, and CNNs. Compared to the other classifiers, the RFC seems to be most robust in case of a strong reduction of data (i.e., the time normalization to 11 data points or the unsupervised data reduction using the selection of time-discrete gait variables). However, the highest performance scores were achieved by SVMs followed by MLPs, RFCs, and CNNs. For gait data the SVM seems to be a powerful machine-learning classifier as often described in the literature (Figueiredo et al., 2018). The MLPs provided only mediocre prediction performances, which could be due to the fact that the total amount of data is simply too small for ANNs (Chau, 2001b; Begg and Kamruzzaman, 2005; Begg et al., 2005; Lai et al., 2008). This impression is reinforced by the even lower prediction performances of the CNNs as "deep" ANN architecture. In addition, the MLPs and CNNs required a lot of computation time for the classification, while the classification based on SVM and RFC was much more timesaving. Based on the presented results, using linear SVMs for the classification of gait data can be recommended. Furthermore, in line with recent research (Slijepcevic et al., 2020), a majority vote could possibly provide an even better classification. However, it should be noted that only a small selection of classifiers and architectures were examined in this analysis.

## CONCLUSION

Based on a systematic comparison, the results provide first domain-specific recommendations for commonly used preprocessing methods prior to classifications using machine learning. However, caution is advised here, as the present findings may be limited to the classification task examined (six-session classification of intra-individual gait patterns) or even to the dataset. Furthermore, the derived recommendations are based exclusively on the prediction scores of the models. Therefore, no information can be obtained about the actual impact of the preprocessing methods and their combinations on the training process and the class representations of the trained models. Overall, it can be concluded that preprocessing has a crucial influence on machine-learning classifications of biomechanical gait data. Nevertheless, further research on this topic is necessary to find out general implications for domain-specific standard procedures.

## DATA AVAILABILITY STATEMENT

The dataset analyzed during the current study is available in the Mendely Data Repository (Burdack et al., 2020) (http://dx.doi.org/10.17632/y55wfcsrhz.1).

## ETHICS STATEMENT

The studies involving human participants were reviewed and approved by the ethical committee of the medical association Rhineland-Palatinate in Mainz (Germany). The participants provided their written informed consent to participate in this study.

## AUTHOR CONTRIBUTIONS

## ACKNOWLEDGMENTS

We would like to thank John Henry Challis for providing his Matlab files to determine the optimal filter cut-off frequency and Djodje Slijepcevic for supporting us with his code for the PCA. This research has been partially funded by the German Federal Ministry of Education and Research (Grant No. 01S18038B).

## SUPPLEMENTARY MATERIAL

The Supplementary Material for this article can be found online at: https://www.frontiersin.org/articles/10.3389/fbioe.2020.00260/full#supplementary-material

**Conflict of Interest:** The authors declare that the research was conducted in the absence of any commercial or financial relationships that could be construed as a potential conflict of interest.